\documentclass[conference]{IEEEtran}
\IEEEoverridecommandlockouts
\usepackage{cite}
\usepackage{amsmath,amssymb,amsfonts}
\usepackage{algorithmic}
\usepackage{graphicx}
\usepackage{textcomp}
\usepackage{xcolor}
\usepackage{hyperref}
\usepackage{lipsum}
\usepackage[para]{footmisc}

\usepackage{booktabs}           % professional-quality tables
\usepackage{multirow}           % tabular cells spanning multiple rows
\usepackage{duckuments}         % sample images

\usepackage{arydshln}
\usepackage{float}

\def\BibTeX{{\rm B\kern-.05em{\sc i\kern-.025em b}\kern-.08em
    T\kern-.1667em\lower.7ex\hbox{E}\kern-.125emX}}
\begin{document}

\title{Multi-Relational Graph Neural Network for Out-of-Domain Link Prediction}
% \author{\IEEEauthorblockN{Anonymous Authors}}
\author{\IEEEauthorblockN{Asma Sattar\IEEEauthorrefmark{1},
Georgios Deligiorgis \IEEEauthorrefmark{2},
Marco Trincavelli\IEEEauthorrefmark{3}, 
Davide Bacciu\IEEEauthorrefmark{4}
}
\IEEEauthorblockA{\IEEEauthorrefmark{1}\IEEEauthorrefmark{4}
\textit{Dipartimento di Informatica}, 
\textit{Universit\unexpanded{á} di Pisa}, 
\text{Pisa, Italy}
}
\IEEEauthorblockA{\IEEEauthorrefmark{2}\IEEEauthorrefmark{3}
\textit{Business Tech, AI, Analytics, and Data}, 
\textit{H\&M Group}, 
Stockholm, Sweden
}

\IEEEauthorblockA{
\IEEEauthorrefmark{1}asma.sattar@phd.unipi.it,
\IEEEauthorrefmark{2}georgios.deligiorgis@hm.com,
\IEEEauthorrefmark{3}marco.trincavelli@hm.com,
\IEEEauthorrefmark{4}bacciu@di.unipi.it}
% \\[-3.0ex]
}

% \author{
% \IEEEauthorblockN{Asma Sattar }
% \IEEEauthorblockA{\textit{Dipartimento di Informatica} \\
% \textit{Universit\unexpanded{á} di Pisa}\\
% Pisa, Italy \\
% asma.sattar@phd.unipi.it}
% \and
% \IEEEauthorblockN{Georgios Deligiorgis and Marco Trincavelli}
% \IEEEauthorblockA{\textit{Business Tech, AI, Analytics, and Data} \\
% \textit{H\&M Group}\\
% Stockholm, Sweden \\
% \{georgios.deligiorgis,marco.trincavelli\}@hm.com}
% % \and
% % \IEEEauthorblockN{Marco Trincavelli}
% % \IEEEauthorblockA{\textit{Business Tech, AI, Analytics, and Data} \\
% % \textit{H\&M Group}\\
% % Stockholm, Sweden\\
% % marco.trincavelli@hm.com}
% \and
% \IEEEauthorblockN{Davide Bacciu}
% \IEEEauthorblockA{\textit{Dipartimento di Informatica} \\
% \textit{Universit\unexpanded{á} di Pisa}\\
% Pisa, Italy \\
% bacciu@di.unipi.it}}
% % \and
% % \IEEEauthorblockN{5\textsuperscript{th} Given Name Surname}
% % \IEEEauthorblockA{\textit{dept. name of organization (of Aff.)} \\
% % \textit{name of organization (of Aff.)}\\
% % City, Country \\
% % email address or ORCID}
% % \and
% % \IEEEauthorblockN{6\textsuperscript{th} Given Name Surname}
% % \IEEEauthorblockA{\textit{dept. name of organization (of Aff.)} \\
% % \textit{name of organization (of Aff.)}\\
% % City, Country \\
% % email address or ORCID}
% }

\maketitle

\begin{abstract}
Dynamic multi-relational graphs are an expressive relational representation for data enclosing entities and relations of different types, and where relationships are allowed to vary in time. 
Addressing predictive tasks over such data requires the ability to find structure embeddings that capture the diversity of the relationships involved, as well as their dynamic evolution. 
In this work, we establish a novel class of challenging tasks for dynamic multi-relational graphs involving out-of-domain link prediction, where the relationship being predicted is not available in the input graph. 
We then introduce a novel Graph Neural Network model, named GOOD, designed specifically to tackle the out-of-domain generalization problem. 
GOOD introduces a novel design concept for multi-relation embedding aggregation, based on the idea that \textit{good} representations are such when it is possible to disentangle the mixing proportions of the different relational embeddings that have produced it. 
We also propose five benchmarks based on two retail domains, where we show that GOOD can effectively generalize predictions out of known relationship types and achieve state-of-the-art results.
Most importantly, we provide insights into problems where out-of-domain prediction might be preferred to an in-domain formulation, that is, where the relationship to be predicted has very few positive examples.
\end{abstract}


\section{Introduction}
\label{Introduction}
Graphs are popular abstractions that represent complex compound data made up of several entities, and nodes, connected by relationships, and represented by graph edges. Graph Neural Networks (GNNs) \cite{scarselli2008graph,zhang2018link,zhou2020graph} provide an effective means of mapping such discrete combinatorial objects and their composing substructures into a numerical representation that facilitates their use in predictive machine learning tasks, with a variety of applications spanning computer vision \cite{wang2018zero}, natural language processing \cite{yao2019graph}, recommendation systems \cite{sattar2021context}. The key intuition underlying most of such GNN models is the computation of a node vectorial representation (embedding) by an iterative process of aggregation of messages from nearby nodes and propagation of messages to the same neighbors.

Node relationships, and hence graph edges, are central to such a message-passing mechanism. In their simplest and most popular form, graphs are typically restricted to edges that represent a single static relationship (i.e., not evolving in time). Real-world processes, however, often call for richer representation capabilities that can accommodate nodes of different types and edges representing different kinds of relationship (e.g., friendships vs. co-working vs. family). The resulting networks are often referred to as multi-dimensional or multi-relational graphs. 
Previous work has dealt with these structures mostly by transforming, more or less explicitly, a single multi-relational graph into several graphs, one for each node/edge type, producing embeddings independently for each relation \cite{chen2021msgcn,schlichtkrull2018modeling}. This approach treats the different relationships in isolation, ignoring their cross-dependencies and limiting the quality of the representation learned by the GNN and its predictive power. To exemplify this aspect, let us consider a scenario comprising products sold in four countries (A, B, C, D), where the former are nodes, and each country defines a different relationship type. Here, the edges represent ``bought-together'' and ``style-together'' relationships. Now assume that A is a neighboring country to B, while C is a neighbor to D. Knowing that two products have been purchased/styled in country A but not in countries C and D can affect the prediction of future purchases in country B, allowing us to take into consideration the different behaviors with different contributing factors. 

The above limitation of multi-relational graph models in the literature pairs with an additional and most crucial one: lack of consideration for out-of-domain relationship prediction. Work in the literature determines the existence of a relationship between two nodes (i.e., the presence of an edge) under the assumption that the input multi-relational graph already contains examples of links of the same type (in-domain scenario) \cite{sankar2020dysat}. Recalling our example above, we would like to be able to learn to predict the edges of an out-of-domain country E by leveraging a multi-relational graph encoding only relationships of type A-D. 

The work described in this paper is, to the extent of our knowledge, the first to address the problem of link prediction in multi-relational graph data in such an out-of-domain scenario. To make the problem more challenging and general, we also release the constraint on static graphs, 
%asma%%allowing (a subset of) relationships to have edges that vary dynamically in time. 
allowing (a subset of) relationships to have edges that vary dynamically with respect to context/relation. 
%asma%%
In the remainder of this paper, we introduce GOOD, a multi-relational \textbf{G}NN for \textbf{O}ut-\textbf{O}f-\textbf{D}omain prediction problems. The key methodological contribution of GOOD, which allows the transfer of knowledge to out-of-domain relationships, lies in the use of a novel relationship aggregation component paired with a disentanglement loss. The key intuition is that a single multi-relational graph representation can be straightforwardly obtained by a weighted aggregation of the contributions from each relationship type; however, in order for this representation to be ``good'', the model should be able to reconstruct the mixing proportions of every single relationship from the aggregated embedding.  

\begin{figure}[tb]
    \centering
    \includegraphics[height=4.2cm]{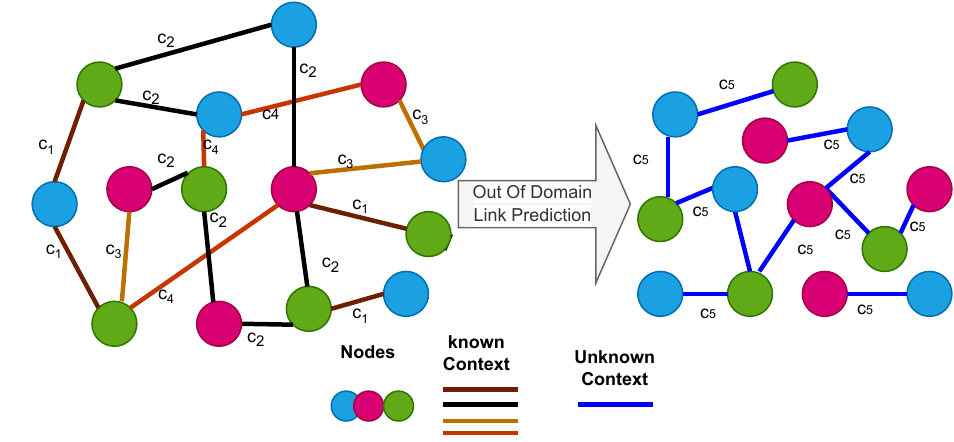}
    \caption{A multi-relational graph with nodes of different types connected with known relations (left), that are used to predict an unknown relationship (right) in a link prediction task.}
    \label{fig:hetero_graph}
\end{figure}

Fig.~\ref{fig:hetero_graph} provides a pictorial intuition of our out-of-domain task: given information about known relations ($C'=\{c_1,c_2.c_3,c_4\}$) between entities (on the left), the task is to predict an unseen relation ($c_5 \text{ or } C'+1$). 
%asma%
% between the same entities (on the right). 
For the sake of compactness, we will refer to the setting in Fig.~\ref{fig:hetero_graph} as Multi-Input Single-Output (MISO), in contrast to single-Input Single-Output (SISO) in-domain learning.
We also assume that we work with a finite number of relationships, which we also refer to as contexts in the rest of the paper. 

The key contributions of this work are as follows.
\begin{itemize}
    \item We formalize the novel problem of out-of-domain link prediction in multi-relational graphs and we introduce associated benchmarks from two retail domains. 
    \item We propose a novel GNN model named GOOD, capable of addressing the MISO learning problem in multi-relational graphs. To the best of our knowledge, this work is the first to extend multi-relational GNNs to learn from known domain relations and predict out-of-domain contexts.
    \item We introduce the idea of regularizing a learning model to be proficient in separating aggregated embeddings into their mixing coefficients, as a driver for obtaining more effective representations for out-of-domain generalization.
    \item We provide an empirical validation of the effectiveness of our approach in five benchmark tasks, showing that GOOD can outperform models from the related literature in both the MISO and SISO settings.
\end{itemize}

\section{Related Work}
Several models exist that tackle the problem of learning representations from homogeneous graphs \cite{hamilton2017inductive,kipf2017semi,velivckovic2017graph}, which have rarely found application to multi-typed edges by straightforward extensions of the model that leverage type-dependent parameterizations. A discussion of these models is beyond the scope of this work, as they do not deepen the aspect of aggregation of multiple relationships into a single embedding, and they have little generality and portability to the MISO setting: the interested reader can refer to popular surveys for further details \cite{bacciu2020gentle, cai2018comprehensive}.  

% Recently, work has started to deal specifically with representation learning on multi-graphs \cite{ma2018multi,qu2017attention}. Earlier work has been developed in the context of knowledge graphs (KG), where different relations are mapped to different aggregation operators working on the KG concept embedding. The Neural Tensor Network (NTN) \cite{socher2013reasoning}, for example, defines each relationship as a bilinear tensor operator followed by a linear matrix operator in concept embeddings. The TransE model \cite{bordes2013translating} represents each relation as a single vector that interacts linearly with the concept vectors. Random walk-based approaches have also been extended from the original skip-gram model to leverage meta-paths for multi-relational data, e.g., MAGNN \cite{fu2020magnn}, HIN2Vec \cite{fu2017hin2vec} and HAN \cite{wang2019heterogeneous}. In addition, in this case, applications are limited to the SISO setting alone, and their high computational requirements affect their scalability to realistic settings.

Recently, work has started to deal specifically with representation learning on multi-relational graphs \cite{qu2017attention,potluru2020deeplex,fang2022pf}. Earlier work has been developed in the context of knowledge graphs (KG), where different relations are mapped to different aggregation operators working on the KG concept embedding. For example, neural tensor network (NTN) \cite{socher2013reasoning} defines each relationship as a bilinear tensor operator, TransE \cite{bordes2013translating} represents each relation as a single vector and random walk-based approaches, e.g., MAGNN \cite{fu2020magnn}, HIN2Vec \cite{fu2017hin2vec} and HAN \cite{wang2019heterogeneous} leveraging meta-paths for multi-relational data. In addition, in this case, applications are limited to the SISO setting alone, and their high computational requirements affect their scalability to realistic settings.

Schlichtkrull et al. \cite{schlichtkrull2018modeling} were among the first to extend the popular graph convolution approach \cite{kipf2017semi} to multi-relational graphs. 
% MSGCN \cite{chen2021msgcn} and 
M2GRL \cite{wang2020m2grl} learns a representation of the nodes of each subrelation graph separately and then aggregates the final representation for a downstream node classification task.
% Similarly, PTE \cite{tang2015pte}, which is an extension of LINE \cite{tang2015line}, splits the input multi-graph into many homogeneous and bipartite networks. 
 The recent MB-GMN  model \cite{xia2021graph} exploits multi-relationships/behaviors in heterogeneous graphs and learns the type-dependent representation of behavior using GNN to capture personalized high-order collaborative effects. This model is not capable of out-of-domain link prediction. Recently, research has introduced models for discrete dynamic graphs \cite{liu2020exploiting,sankar2020dysat,chen2021msgcn}. DSRS\cite{liu2020exploiting} learns the dynamic social influence of users on their preferences for items by employing structural and temporal attention. 
 % Similarly, the DySAT \cite{sankar2020dysat} consists of structural and dynamic temporal blocks. The structural block applies GAT \cite{velickovic2018pietro} on T discrete graphs to extract the node’s local neighborhood features along with self-attention aggregation. 
 The models proposed in \cite{hang2022lightweight,chen2021msgcn} provide the setting to learn the representation of new nodes entering the discrete dynamic network. In MSGCN \cite{chen2021msgcn}, the representation of the new node is generated using vector representations of neighboring nodes from multiple discrete relational graphs (known contexts).
 % Because nodes typically have close relationships with their neighbors, these known nodes can be utilized to obtain the representations of new nodes.

In general, work in the literature is limited to an in-domain setting, and little effort has been put into investigating general forms of aggregation across multiple contexts. As a result of this, the aggregation functions are typically chosen to optimize an in-domain performance metric (e.g., node classification accuracy), and the models are not flexible enough to account for out-of-domain predictive tasks. We believe that the latter is a challenging problem which nevertheless has high potential in settings characterized by the richness in terms of the number of types of relationship available, but where the number of positive examples for each relationship type is relatively low. In this particular setting, we expect to gain the most from a truly integrative approach, such as GOOD, operating across different relationships to obtain high-quality and general embeddings of the structures.

\section{Problem Definition}
\begin{figure*}[tp]
    \centering
    \includegraphics[width=13.5cm,height=7.5cm]{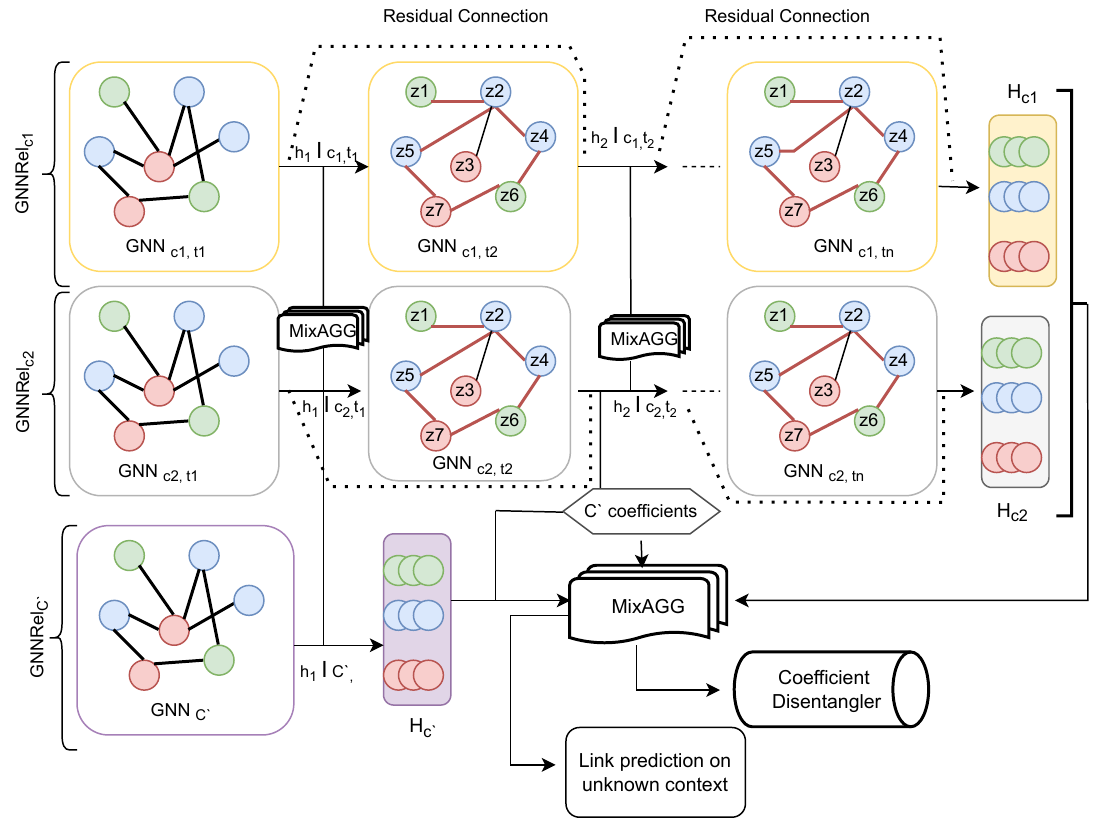}
    \caption{High-level architecture diagram. From top to bottom: each row represents node embeddings generated with respect to different contexts. From left to right: each row represents the effect of a sub-context inside a main context. Dashed lines represent the residual connections. MixAGG denotes the aggregation block.}
    \label{fig:arch_diagram}
\end{figure*}
%asma%%
A multi-relational graph $\mathcal{G(V, E)}$ consists of $M$ types of nodes and $C$ types of edges (Fig.~\ref{fig:hetero_graph}). Let $\mathcal{V} = \{\mathcal{V}_1, \mathcal{V}_2, ..., \mathcal{V}_M \}$ represent a set of $M$ types of nodes, where $|\mathcal{V}|$ represents the total number of nodes of all types and $|\mathcal{V}_i|$ is the total number of nodes in the set $i$.

The nodes in the graph are connected with $C$ types of edges. When considering dynamic graphs in which edges can change over time, edges can be represented as $\mathcal{E}= \big\{ \mathcal{E}_{c,t} | c: \{c_1, c_2, ..., c_{C} \}, t: \{ 1, 2, ..., T - 1, T \} \vee \emptyset \big\} $ where $\mathcal{E}_{c,t}$ is a set of links of type $c$ at time $t$. The number of contextual relations available is denoted by $C$, while $T$ is the total number of time steps. The number of edges in a given context and time is denoted by $|\mathcal{E}_{c, t}|$. 

%asma%%
The out-of-domain task for discrete context-based dynamic graphs refers to the problem of learning from a set of known context relational graph snapshots $\mathcal{G}: \{\mathcal{G}_{c_1}, \mathcal{G}_{c_2}, \cdots, \mathcal{G}_{C'} \}$ while predicting unseen relationships for context in $\mathcal{G}_{C'+1}$. Where context explicitly is any information that can be utilized to characterize and interpret the situation in which nodes interact with each other. As shown in Fig.~\ref{fig:arch_diagram}, inside the GNNREl block, we can have different nodes and edges under different contexts, which can refer to different time stamps, locations, etc. More formally, this can be drafted as

% Out-of-domain refers to the problem of learning in a set of known contexts for a period of time $t=1,\dots,T-1$ while predicting unseen relationships and the time step $t=T$. As shown in Fig.~\ref{fig:arch_diagram}, inside the GNNREl block, we can have different nodes and edges under different contexts, which refer to different time stamps. More formally, this can be drafted as
% \begin{center}
% %George%%
% $\mathcal{G}_{c_1}(V_{c_1},E_{c_1}),\dots,\mathcal{G}_{C'}(V_{C'},\mathcal{E}_{C',t=1}),\dots,
% \mathcal{G}_{C'}(V_{C'}, \mathcal{E}_{C',t=T-1})\xrightarrow[\text{domain}]{\text{out of}}\mathcal{G}_{C'+1}(V_{C'+1},E_{C'+1,t=T}),  \dots ,\mathcal{G}_{C}(V_{C},\mathcal{E}_{C})$

\begin{center}

$\mathcal{G}_{c}(V_{c},E_{c}),\dots,
\mathcal{G}_{C',T-1}(V_{C'}, \mathcal{E}_{C',T-1})\xrightarrow[\text{domain}]{\text{out of}}\mathcal{G}_{C'+1,T}(V_{C'+1},E_{C'+1,T}),  \dots ,\mathcal{G}_{C}(V_{C},\mathcal{E}_{C})$
\end{center}
% %asma%%
% $\mathcal{G}_{c_1}(V_{c_1},E_{c_1}),\dots,
% \mathcal{G}_{C',t=1,\dots,T-1}(V_{C',t=1,\dots,t=T-1}, \mathcal{E}_{C',t=1,\dots T-1})\xrightarrow[\text{domain}]{\text{out of}}\mathcal{G}_{C'+1,t=T}(V_{C'+1,t=T},E_{C'+1,t=T}),  \dots ,\mathcal{G}_{C}(V_{C},\mathcal{E}_{C})$
% \end{center}

where $c_{i=1, \dots, C'}$ represent the $C'$ known contexts, $c_{j=C'+ 1, \dots, C}$ are the $C - C'$ unknown target contexts and $E_{C',t}, E_{C}$ are the time dependent and independent contexts, respectively. The arrow denotes the structured transduction between the multi-relation graph in input, defined over $C'$ relationships, and the multi-relation graph in output, defined on a disjoint set of target contexts.

Given this problem setting, we focus mainly on generating strong representations of the nodes conditioned on different contexts/relations\footnote{In the paper we use the terms context, relation, and contextual relation interchangeably when denoting the type of the graph edges. \nopagebreak}.
To achieve this, we build a multi-objective model that learns to generate strong embedding representations for known contexts, but at the same time promotes an efficient combination of representations coming from several contexts in a self-supervised manner. In other words, our recommendation problem is cast as the task of predicting the existence of an unknown contextual relation among nodes, given the relations among the nodes present in different contexts.

\section{Modeling Multi-Relational embeddings for Multi-Relational Graph}
\label{sec:modeling}
This section introduces our graph neural network for Out-Of-Domain (GOOD) link prediction in discrete dynamic multi-relational graphs. The high-level architecture diagram of the proposed model is shown in Fig.~\ref{fig:arch_diagram}. 
The model consists of two key components, the GNN multi-relation embedded (GNNRel) and the Mixing-AGGregator (MixAGG).
\begin{description}
    \item \textbf{GNNRel}: This component constructs an independent representation of the nodes for each relation. 
 As illustrated in Fig.~\ref{fig:arch_diagram}, some relations have a time dependency, while other relations are connected with respect to the time-independent context.
 %asma  independent of time. 
 For time-dependent contexts, we constructed a different graph for each time step. Therefore, each graph considered by the GNNRel module is a homogeneous one considering only one context at a given time step.
The GNNRel is implemented using a Graph Convolutional Network (GCN) \cite{kipf2017semi} and is responsible for generating structure embeddings specialized for single contexts. 
\item \textbf{MixAGG}: The MixAGG component builds an aggregated representation based on the embeddings produced by the GNNRel module. On a high level, we aggregate the contexts using mixing coefficients $Q=\{q_1, q_2, ..., q_{C'}\}$, where $C'$ is equal to the number of known contexts such as $1 \leq C' \leq C, C' \in \mathbb{N^*}$ and $q_i$ has the properties of a probability mass function, that is, $0 \leq q_i, \sum q_i = 1, q_i \in \mathbb{R}$. The mixing coefficients are set in different ways depending on the variation of the model, and the aggregation can be performed using several alternative methods, as discussed in detail in Section \ref{sec:GOOD}.
% which are discussed in detail in Section \ref{sec:AGG}
% The mixing coefficients are set in different ways depending on the variation of the model, as outlined in Section \ref{sec:GOOD} and the aggregation can be performed using several alternative methods, which are discussed in detail in Section \ref{sec:AGG}.
\end{description}

The final embedding vectors are then fed to two Multi-Layer Perceptrons (MLP). The first is responsible for the link prediction task in the unknown domain, and the second is responsible for predicting the mixing coefficients used during the aggregation step. The disentanglement of mixing coefficients through the Coefficient Disentangler (CD) mimics the classic problem of Blind Source Separation \cite{hyvarinen2000independent}, when one wants to discriminate between different sources, compensating for the perceived compound signal.
% (predicting the coefficients used to aggregate the contexts)
% To achieve multi-objectives, here we have another feed-forward neural network that focuses on retrieving the probabilistic contribution of each relation which is used during aggregation. 
We argue that our architecture helps generalization leveraging the information contained in out-of-domain contextual relations, since it pushes the model to learn how to combine and generate strong embedding vectors by conditioning on different contextual relations mixed in the final embedding with coefficients that can even be learned from the data.

In our work, we propose a novel way in which two or more contexts can be combined with any normalized coefficient in a continuous space and still perform adequately in the given target context. 
The goal is to use a different set of coefficients (different contributions) to generate embedding representations, which could be useful for other use cases, which share some common base information with the ones used for training. 

Next, we explain in more detail the variants of our proposed model, which are GOOD, GOOD$_{LC}$ and GOOD$_{LC}^+$ (Section \ref{sec:GOOD}). Subsequently, we will define several possible aggregation methods for the MixAGG block (same section). 
% (Section \ref{sec:AGG}). 
The loss function of a GOOD model is discussed in Section \ref{sec:loss}.
\subsection{Model Architecture}
\label{sec:GOOD}
For the generation of nodes' embeddings, GOOD is using the GNNRel block by treating each context as independent of the other contexts, meaning that each context at a specific time step is a homogeneous graph.
% We treat each relation as independent to the other contexts and the time part is taken into consideration as a sequence. 
On each homogeneous graph, we apply the following sequential layers:
\begin{equation}
    f_{c, t}(\mathbf{X}, \mathbf{A}_{c, t}) = BN \Bigg( DP \Big( ReLU \big( GCN(\mathbf{X}, \mathbf{A}_{c, t}) \big) \Big) \Bigg)
    \label{eq:gnn_rel}
\end{equation}
where ReLU \cite{agarap2018deep} is the Rectified Linear Unit activation function, DP and BN are the Dropout \cite{srivastava2014dropout} and Batch Normalization \cite{ioffe2015batch} layers, respectively. 
The $\mathbf{X}$ are the input embeddings and $\mathbf{A}_{c, t}$ is the adjacency matrix for context $c$ at time $t$ that encodes the edges $\mathcal{E}_{c, t}$. 
Sequential layers in \eqref{eq:gnn_rel} compose the subblocks, and all of these subblocks form the GNNRel model. 
At each time step, we can apply more than one $f_{c,t}$ based on the number of $k-$hop neighbors that we are interested in visiting.
The GNN-Rel model is then formulated as follows:
% \[
%     \mathbf{H}_{c, t_0} = \mathbf{X}
% \]
% \begin{equation}
%     \mathbf{H}_{c, t_1} = f_{c, t_1} ( \mathbf{H}_{c, t_0} \mathbf{A}_{c, t_1} ) 
%     \label{eq-time-in-dependent-context}
% \end{equation}
% \[
%     \mathbf{H}_{c, t_2} = f_{c, t_2} ( \mathbf{H}_{c, t_1} \mathbf{A}_{c, t_2} ) + M_{c, t_2}(\mathbf{H}_{c, t_0}) 
% \]
% $$\cdots$$
\begin{equation}
    \mathbf{H}_{c, t_{T-1}} =  f_{c, t_{T-1}} ( \mathbf{H}_{c, t_{T - 2}} \mathbf{A}_{c, t_{T-1}} ) + M_{c, t_{T-1}}(\mathbf{H}_{c, t_{T-3}}) 
\label{eq-time-context} \end{equation} 
where $\mathbf{H}_{c, t_0} = \mathbf{X}$, $M_{c, t}$ is a matrix that projects the input embedding vectors of the previous layer onto the current if and only if the dimensions of the embeddings of two consecutive time steps are different; otherwise $M_{c, t}$ is equal to the identity matrix. Using the input embeddings of the previous layer as input to the current layer is known as residual connections (skip connections), which are illustrated with dashed lines in Fig.~\ref{fig:arch_diagram}. For a time-independent context, $f_{c}$ has to be applied once in \eqref{eq-time-context}. Moreover, the residual connections computations in \eqref{eq-time-context} are skipped, since there are no historical time steps. 

Finally, the nodes' embeddings for all relations are aggregated as follows:
\begin{equation}
    \mathbf{H} = \text{MixAGG} \Big( ( \mathbf{H}_{c_1}, q_{c_1}), (\mathbf{H}_{c_2},q_{c_2}), \cdots, (\mathbf{H}_{c_{C'}}, q_{c_{C'}}) \Big)
\end{equation}
where $q_{c_i}$ is the coefficient that controls how much each context contributes. The coefficients $q_{c_i}$ are set in different ways in the three different model variations that we propose:
\begin{enumerate}
    \item \textbf{GOOD}: During training, it uses random mixing coefficients sampled at the beginning of each epoch and a CD that identifies which random coefficients are used per context. We sample the coefficients from a Dirichelet distribution of order $C'$ with concentration parameters sampled from a uniform distribution $[0,1)$, ensuring that they are all greater than or equal to zero. During inference, the mixing coefficients are set to $q_i = 1 / C'$, which is the expected value of the concentration parameters of the Dirichelet distribution. 
    \item \textbf{GOOD$_{\text{LC}}$}: This model learns the mixing coefficients during training. Since the learned coefficients are not guaranteed to sum up to one, they are normalized once the training is completed. Moreover, since the mixing coefficients are learned, the CD module is not used in this model. At inference time, the model uses the normalized mixing coefficients obtained in training.
    \item \textbf{GOOD$_{\text{LC}}^+$}: This configuration uses the normalized mixing coefficients learned by the GOOD$_{\text{LC}}$ in the inference phase for a GOOD model trained with random mixing coefficients. The rationale behind this choice is to decouple the learning of the mixing coefficients from the learning of the other components of the model. 
\end{enumerate}

% \label{sec:AGG}
We have evaluated several aggregator functions ($Agg_{sum}$, $Agg_{stack}$, $Agg_{Dsum}$, $Agg_{Dtack}$) for the MixAGG block, where $Agg_{sum}$ and $Agg_{stack}$ are sum and stack operations on all node embeddings weighted by their respective $q_{i}$ (contribution w.r.t. $i^{th}$ relation). Whereas $Agg_{Dsum}$ and $Agg_{Dtack}$ also multiply the weighted/unweighted node's degree with respect to relation along with their respective $q_{i}$.
The choice of the best aggregation function is treated as a hyperparameter and is optimized through model selection.

\subsection{Loss Function}
\label{sec:loss}
GOOD features a multi-objective loss function:
\begin{equation}
    L = L_{link} + L_{Q} 
\end{equation}
where the first component $L_{link}$ aims to maximize link prediction performance for the target context, while the second component $L_{Q}$ learns mixing coefficients so that the model generalizes well across several contexts. The two components of the loss can be seen as two downstream tasks that will be solved using the embedding vectors obtained by the models that aggregate several contexts. In the following subsections, we describe in detail the two components of the loss function.

\subsubsection{Link Prediction Objective}
\label{sec:link_prediction_objective}
The link prediction objective ensures that the embedding obtained by merging the various contexts has a strong ability to predict what would happen in the next time step for the target context. Therefore, we have a link predictor for each target context. In case the target context is time-dependent, we predict the links at time $t=T$.
% which will try to predict the links at time $t=T$ . 
The link prediction tasks are formulated as binary classification problems with BCE loss.
%, i.e., with a Binary Cross Entropy (BCE) loss function: 
\begin{equation}
    L_{link} = \frac{ \sum_c \Big( \mathbf{Y}_c log_e \big( \sigma (\mathbf   {\hat{Y}_c}) \big) + (1 - \mathbf{Y}_c) log_e \big( 1 - (\mathbf   {\hat{Y}_c}) \big) \Big)}
    {C - C'}
\end{equation}
where $\sigma(\cdot)$ is a sigmoid activation, %function $\sigma(x) = \frac{1}{1+e^{-x}}$, 
$\mathbf {Y_c}$ are the targets, and $\mathbf {\hat{Y}_c}$ are the predictions of the model:
    \begin{equation}
        \mathbf{\hat{Y}} = \{ \mathbf{\hat{Y}_c} = g_L(\mathbf{H}_{src} \odot \mathbf{H}_{dst}) \quad c: c_1, \dots, c_{C'} \}.
    \end{equation}
The term $g_L(\cdot)$ is the output of the MLP used for link prediction:
% \lipsum

    $
        g_L(\mathbf{H}_{src} \odot \mathbf{H}_{dst}) = \sigma \Bigg( \mathbf{W}_L \text{DP} \Big( \text{ReLU} \big( \text{BN} ( \mathbf{W}_l (\mathbf{H}_{\text{src}} \odot \mathbf{H}_{\text{dst}}) + \mathbf{b}_l ) \big) \Big) + \mathbf{b}_L \Bigg)
    $
    
where $\odot$ is the element-wise notation, $\mathbf{H}_{\text{src}}, \mathbf{H}_{\text{dst}}$ are the embeddings of the source and destination nodes, respectively, Terms $l$ and $L$ denote the $l-\text{th}$ and the last layer of the MLP, respectively, %$\sigma(\cdot)$ is the Sigmoid activation function, 
and $\mathbf{W}_l, \mathbf{W}_L, \mathbf{b}_l, \mathbf{b}_L$ are the learnable weights and biases. 
%The Sigmoid activation function will help solve the classification link prediction task by predicting the existence of a link. 
A common practice to train GNNs for link prediction is to provide a balanced number of positive and negative links as targets. We experimented with different techniques to sample negative links as described in Appendix \ref{sec.implem}.

\subsubsection{Coefficient Disentanglement Objective}
\label{sec:coefficient_disentanglement_objective}
The Coefficient Disentanglement objective enforces the model to learn (reconstruct) the mixing coefficients that are used to calculate the merged embedding representations by aggregating the context-specific ones.
% The Coefficient Disentanglement objective serves the learning of the mixing coefficients to calculate the merged embedding representations by aggregating the context-specific ones. 
This can be seen as a regularization term, which enables the model to generalize better by aggregating information from different contexts. This amounts to a regression task learned by Mean Squared Log Error (MSLE): 
    \begin{equation}
        L_{Q} = \frac{1}{C'} \sum_c \Big( log_e \big( 1 + \mathbf{q}_c \big) - log_e \big( 1 + \mathbf{\hat{q}}_c \big) \Big)^2
    \end{equation}
where $\mathbf{q_c}$ is the vector of the target coefficients and $\mathbf{\hat{q}_c}$ is the vector of the predicted coefficients for context $\mathbf{c}$ computed as
    \begin{equation}
        \mathbf{\hat{q}} = \text{SoftMax} \big( g_Q(\mathbf{H}) \big)
    \end{equation}
where $\mathbf{H}$ are the final embedding vectors and
$g_Q(\cdot)$ is the output of the MLP used for disentangling coefficients, that is:
\begin{equation}
    g_Q (\mathbf{H}) = \mathbf{W}_Q \text{DP} \Big( \text{ReLU} \big( \text{BN} ( \mathbf{W}_q \mathbf{H} + \mathbf{b}_q) \big) \Big) + \mathbf{b}_Q.
\end{equation}
The terms $\mathbf{W}_q, \mathbf{W}_Q, \mathbf{b}_q, \mathbf{b}_Q$ are the learnable weights and biases.
%The SoftMax enforces the output to be normalized and plays a significant role in solving the regression task of disentangling the coefficients.
It should be noted that for the GOOD$_{\text{LC}}$ variant, where the mixing coefficients embedded in the model are learned during training, the objective of the coefficient disentangler $L_Q$ is deactivated. The impact of CD is studied in Appendix \ref{sec:impact_coef}.

\section{Benchmarks and Experimental Analysis}
\subsection{Datasets}

One of the motivating tasks behind our problem formulation is the need to build a predictor of future co-purchases (products bought in the same basket) of $C - C'$ relations, knowing what has been purchased from $C'$ relations. Coherently with such an objective, we have assembled five benchmark datasets to assess our model, generated in the context of retail data. Three datasets, that is, ST$_1$, ST$_2$, and ST$_3$, are obtained from an %industrial
internally available Seasonal Transaction (ST) dataset provided by a global fashion retailer H\&M.
% \footnote{The name is not shown for preserving anonymity.}.  %H\&M
The other two datasets, CS$_{1}$ and CS$_{2}$, are obtained from publicly available data known as Customer Segmentation \footnote{\url{https://www.kaggle.com/fabiendaniel/customer-segmentation/data}}.
All datasets can be represented through a bidirectional graph, where the nodes represent the products and the links connect the products bought in the same basket. 
Two products can be connected multiple times with respect to different contexts.
More statistical information on the data sets can be found in \autoref{tab:dataset}.
%H\&M is one of the biggest global fashion retailers with high quality data.

The ST$_{1}$, ST$_{2}$ and ST$_{3}$ datasets consist of online transactions from multiple markets and different seasons. The nodes' representations are visual features extracted from a ResNet-50 \cite{he2016deep} pre-trained on ImageNet \cite{deng2009imagenet}. Each dataset is a multi-relational graph that consists of four types of products and twenty-four types of edges (comprising four contexts and six seasons). Three contexts are used as input, and the fourth as the target of the model. For training, we used seasons $1, 2, 3 \xrightarrow{} 4$, for validation, we used seasons $2, 3, 4 \xrightarrow{} 5$ and for testing, we used seasons $3, 4, 5 \xrightarrow{} 6$. %The density of the test graph is shown in \autoref{tab:dataset}.

\begin{table}[tp]
% \begin{table*}[!ht]
    \caption{Statistical information about external and internal datasets.}
     \label{tab:dataset}
    \centering
    % \tiny
    % \resizebox{1.2\columnwidth}{!}{
\begin{tabular}{lcccc}
\toprule
% Dataset  & 
Dataset  & Type of Nodes & Type of Edges & \texttt{\#}Nodes & \texttt{\#}Edges 
% & \texttt{\#}Test-graph density
\\
\midrule
% \multirow{3}{*}{Seasonal Transactions} & 
ST$_{1}$ & \multirow{3}{*}{4} & \multirow{3}{*}{24} & \multirow{3}{*}{72549} & \multirow{3}{*}{6726113} 
% & 0.000049 
\\
% & 
ST$_{2}$ & & & & 
% & 0.000026
\\
% & 
ST$_{3}$  & &  &  &  
% & 0.000042 
\\
%& ST$_{4}$ & & & & & 0.000114 \\
% \begin{tabular}{@{}c@{}} c1=0.000048  \\ c2=0.000026 \\ c3=0.000114  \\ c4=0.0000422 \end{tabular} \\
% \hline
\hdashline
% \multirow{2}{*}{Customer Segmentation} & 
CS$_{1}$ & \multirow{2}{*}{1} & \multirow{2}{*}{18} & \multirow{2}{*}{3729} & \multirow{2}{*}{3355544} 
% & 0.000007 
\\
% & 
CS$_{2}$ & & & & 
% & 0.000880 
\\
%& UK$_{3}$ & & & & & 0.208421 \\
% \begin{tabular}{@{}c@{}} g1=0.0000074 \\ g2=0.000880 \\ g3=0.20842145 \end{tabular} \\ 
% \hline
  \bottomrule
\end{tabular}
    % }
% \end{table*}
\end{table}
\begin{table*}[htp]
    \caption{Test-set performance comparison of the proposed approach against related models from the literature. The parameters used to achieve these results are listed in Appendix \ref{sec.implem}.}
    \centering
    \label{tab:hm_results}
    \resizebox{\textwidth}{!}{%
    \begin{tabular}{lcccccccccc}
\toprule
\multirow{3}{*}{\textbf{ALGORITHM}} 
% & \multicolumn{8}{c}{\textbf{H\&M}}  
& \multicolumn{2}{c}{\textbf{ST$_1$}}  
& \multicolumn{2}{c}{\textbf{ST$_2$}}  
& \multicolumn{2}{c}{\textbf{ST$_3$}}  
% & \multicolumn{2}{c}{\textbf{$c_4$}} 
& \multicolumn{2}{c}{\textbf{CS$_1$}}  
& \multicolumn{2}{c}{\textbf{CS$_2$}} 
\\
% \cline{2-5}
& \textbf{Accuracy}      
& \textbf{ROC}    
& \textbf{Accuracy}        
& \textbf{ROC} 
& \textbf{Accuracy}      
& \textbf{ROC}    
% & \textbf{Accuracy ($\%$)}        
% & \textbf{ROC-AUC} 
& \textbf{Accuracy}      
& \textbf{ROC}    
& \textbf{Accuracy}      
& \textbf{ROC}    
\\

& 
($\%$) & \textbf{AUC} &
($\%$) & \textbf{AUC} &
($\%$) & \textbf{AUC} &
($\%$) & \textbf{AUC} &
($\%$) & \textbf{AUC}\\

\midrule
GOOD & 
% C1 (NL)
71.28 & \textbf{0.6340} &
% C2 (IT)
65.09 & \textbf{0.5963} &
% C3 (SE)	
62.54 & \textbf{0.6484} &
%%%%%%%%%%uk%%%%%%%%%
% g1
\textbf{64.42} &  \textbf{0.6085} & 
% g2
66.12  & 0.6990  
\\
GOOD$_{\text{LC}}$ &
% C1 (NL)
71.18  & 0.6338 &
% C2 (IT)
59.45 & 0.5924 &
% C3 (SE) 
62.58 & 0.6341 &
%%%%%%%%%%uk%%%%%%%%%
% g1
57.21 & 0.5973 & 
% g2
63.75 & 0.6884\\
GOOD$_{\text{LC}}^+$ & 
% C1 (NL)
\textbf{71.32} & 0.6307 &
% C2 (IT)	
\textbf{66.64} & 0.5935 &
% C3 (SE) 	
\textbf{62.61} & 0.6430 &

%%%%%%%%%%uk%%%%%%%%%
% g1
\textbf{64.42} & \textbf{0.6085}& 
% g2
\textbf{66.14} & \textbf{0.6991} 
\\
\hdashline
GCN \cite{kipf2017semi} & 
% C1 (NL)	
71.12  & 0.6078 &
% C2 (IT)
\textbf{71.21} & 0.5733 &
% C3 (SE)
61.89 & 0.6422 &
%%%%%%%%%%uk%%%%%%%%%
% g1
50.48 & 0.4066& 
% g2
53.64 & 0.6161\\
MSGCN \cite{chen2021msgcn} &
% NL (C1) 
71.07  & 0.5000  &
% C2 (IT)	
61.07 & 0.5590&
% C3 (SE)	
60.39 & 0.5998 &
%%%%%%%%%%uk%%%%%%%%%%%%%%%
% g1
54.33 & 0.5861 & 
% g2
44.39  & 0.4464 
\\
MSGCN-mean \cite{chen2021msgcn} & 
% C1 (NL) 	
71.07 & 0.5352 & 
% C2 (IT)	
70.86 & 0.5954 &
% C3 (SE)
61.22 & 0.6372 &
%%%%%%%%%%uk%%%%%%%%%%%%%%%
% g1
39.42 & 0.3859 & 
% g2
52.69 & 0.6293\\
RGCN \cite{schlichtkrull2018modeling} & 
% C1 (NL) 	
 65.84 & 0.6242 & 
% C2 (IT)	
 64.73 & 0.5955 &
% C3 (SE)
62.58
 %\textbf{62.86} 
 &
 0.6431
 &
%%%%%%%%%%uk%%%%%%%%%%%%%%%
% g1
59.62 & 0.5856 & 
% g2
 61.25
 & 0.6592
\\
MMGCN \cite{wei2019mmgcn} & 
% C1 (NL) 	
35.08 & 0.5009  & 
% C2 (IT)	
32.48 & 0.5025 &
% C3 (SE)
43.19 &  0.5068 &
%%%%%%%%%%uk%%%%%%%%%%%%%%%
% g1
50.01 & 0.6074 & 
% g2
49.80 
 & 0.5699
\\
\bottomrule
    \end{tabular}%
    }
\end{table*} 
\begin{figure}[b]
    \centering
    \includegraphics[height=4.3cm]{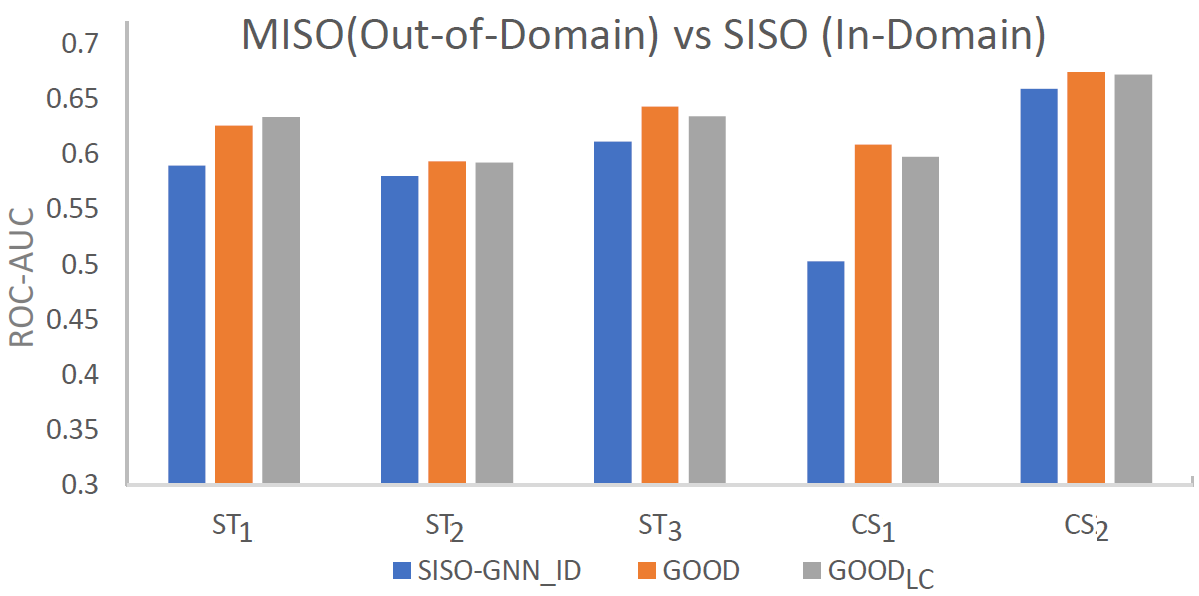}
    \caption{Multi-relational out-of-domain Vs. Single-relation in-domain ROC-AUC results.}
    \label{fig:siso-vs-miso-both}
\end{figure}
The public data sets CS$_{1}$ and CS$_{2}$ consist of online transactions of customers belonging to 37 different countries from a physical store in the UK. 
% The purchases made by ~4000 customers over a period of one year (from 2010-12-01 to 2011-12-09).
Due to the excessive sparsity in the data, we decided to split the contexts in a different way compared to the ST dataset, by grouping countries together under the same context based on their geographical location. Here, the task requires predicting the co-purchases in one group from the purchases that occurred in the other groups. These multi-relational graphs connect one type of product with 16 different types of edges. The initial node embeddings used in the CS datasets are learnable embeddings (for more information, see Section \ref{sec.implem}). 

\subsection{Evaluation Methodology}
\subsubsection {Related Approaches}
We have compared our proposed algorithms GOOD, GOOD$_{\text{LC}}$ and GOOD$_{\text{LC}}^+$ to two families of related works. The first category includes a method that is designed for homogeneous networks.
The second category includes models that are designed for multi-relational graphs. The baseline algorithms used in the empirical evaluation are summarized below.
\begin{itemize}
    \item GCN \cite{kipf2017semi}: A classical semi-supervised GNN (from first category) for aggregating neighborhood information. It is the most referenced method in the GNN literature and the most widely used design in real-world applications.
    \item MSGCN \cite{chen2021msgcn}: MSGCN belongs to the second category; it aggregates the embedding of all relational subgraphs and uses the aggregated embedding (each node has a different weight $w_i$ based on its contribution to the respective relational subgraph) for the node classification task. In our setting, we have used this approach to aggregate the known subgraph relations and predict links for an unknown relational subgraph.
    \item MSGCN-mean \cite{chen2021msgcn}: This is an MSGCN variant that averages all $C'$ relations without taking into account the effect ($w_i$) of the relational graph on each node.
    \item MMGCN \cite{wei2019mmgcn}: builds an independent graph for each context and uses a modified GNN to propagate the information for each graph separately. In its aggregation layer, it learns several preference functions that eventually combine into a single overall node's representation. In the original work, each graph was separated with respect to node features, but here we have a separation on a relation basis (belonging to the second category).
    % This model has been introduced mainly for data sets that contain user-item interactions with different relations among the users and the items. Unlike the original work, we have not used node ID-embedding as they have used for user-item interaction, since the datasets used in this work consist of a single type of nodes, meaning that we have only item-item interactions.
    \item RGCN \cite{schlichtkrull2018modeling}: it uses a relational message passing over the nodes on multi-relational graphs (belongs to the second category). This model performs relation-specific aggregation at the node level without considering the significance of various relations and neighborhoods.
\end{itemize}

\subsection{Performance Comparison}
\label{sec:performance_comparison}
For comparing our models, we mainly used two metrics: Accuracy and Receiver Operating Characteristic Area Under Curve (ROC-AUC) \cite{fawcett2006introduction}. \autoref{tab:hm_results} shows the performance of GOOD and related models in five data sets. GOOD outperforms all models considered from the literature in terms of the ROC-AUC metric.
The inability of MMGCN to generalize on out-of-domain predictions makes it perform the worst compared to our GOOD.
Our model uses both MixAGG with random coefficients and a CD, and generalizes better to out-of-domain tasks, compared to models without MixAGG (i.e., RGCN) and CD (i.e., MSGCN and MSGCN mean).

The purpose of solving an out-of-domain task is two-fold. First, we want to predict an unknown context and, second, to benefit from the knowledge of the other contexts. To evaluate the latest, we compare our model with the same target $c_k$ once using inputs $c_{1}$ to $c_{k-1}$ (out-of-domain) and once only using $c_k$ (in-domain).
Intuitively, the best performance should be achieved when we learn and predict in the same domain.

To this end, we also analyze how the performance of our MISO model in the out-of-domain task compares with the SISO in-domain task. For this purpose, we have used a single GNNRel block per time $t$ in our model (Fig.~\ref{fig:arch_diagram}) to learn and predict in the same domain; we call this model Single Input Single Output in-domain (SISO-GNN-ID). 

In Fig.~\ref{fig:siso-vs-miso-both}, we can see that our model GOOD and GOOD$_{\text{LC}}$ (for the MISO task) is more confident compared to SISO-GNN-ID.
The better ROC-AUC result of GOOD and GOOD$_{\text{LC}}$ is a very positive sign for believing in our model for out-of-domain link prediction.
The results show that multiple relevant relations that are out-of-domain are helpful in generating strong embeddings, which are capable of achieving a performance that is at least as good as the performance of in-domain link prediction. 

\section{Conclusion and Future Work}
We formalized and studied the problem of out-of-domain link prediction on dynamic multi-relational graphs. We argued that existing literature works leveraging graph neural networks for multi-relational graphs are only designed for in-domain tasks, and they are not capable to effectively generalize to out-of-domain problems. To address such a limitation, we introduced a novel GNN approach, GOOD, specifically designed for out-of-domain link prediction tasks with dynamically evolving multi-relationship networks. 
We have discussed and empirically analyzed the performance of different variants of our model, with the help of newly introduced benchmarks in the field of retail product recommendation. 
Among the GOOD variants, it is highlighted through the experimental analysis that we performed that the GOOD$_{LC}$ model is the one that generalizes the least.
However, we have also witnessed the advantage of GOOD$_{LC}^+$, which uses a hybrid strategy utilizing the best-normalized coefficients learned by GOOD$_{LC}$, with the static mixing coefficients of vanilla GOOD.

Our experiments reveal that GOOD provides strong and general embeddings for multi-relational graphs by leveraging the disentanglement mechanism that makes the model proficient in guessing the mixing coefficients used in the multi-relation embedding aggregation step. Such embeddings also seem promising for downstream out-of-domain link prediction tasks
%asma
for dynamic graphs.
The applications of the proposed model are not limited to recommendation systems, but could also include the discovery/repurposing of therapies from protein-drug interaction networks and, in general, any problem that can be formalized using heterogeneous graphs. 

In the future, we plan to investigate a scalable meta-path-based approach for our multi-relational graph setting, aiming to produce stronger and more generalized embeddings, which can improve performance on the out-of-domain link prediction task.
% for the same downstream task. 
Moreover, we also plan to investigate an attention-based integration mechanism for multiple relations and nodes and to evaluate our model on new industrially relevant problems.
%Last but not least we would like to investigate the affect of the model on real-day applications. 

\section{Implementation Details} \label{sec.implem}
Our implementation\footnote{\url{https://github.com/asmaAdil/GOOD}} is based on PyTorch \cite{paszke2019pytorch} and PyTorch Lightning\footnote{\url{http://pytorchlightning.ai}}. We also used the Deep Graph Library (DGL) \cite{wang2019deep} to manipulate the graphs and for the standard implementation of GCN layers \cite{kipf2017semi}. 
Most graphs used in GNNs do not provide ground truth negative edges, but the ST${_1}$, ST${_2}$ and ST${_3}$ datasets contain ground truth negative edges, which are based on returned products. During the training part, the negative edges were sampled in different ways based on their frequency (edge weight) using a multinomial distribution, a uniform random distribution (samples from the ground-truth negative edges), a uniform random distribution (samples from all non-positive edges) or a ratio of all the aforementioned.
% Our implementation is based on PyTorch \cite{paszke2019pytorch} and PyTorch Lightning\footnote{\url{http://pytorchlightning.ai}}. We also used the Deep Graph Library (DGL) \cite{wang2019deep} to manipulate the graphs and for the standard implementation of GCN layers \cite{kipf2017semi}. % and Graph SAmple and aggreGatE (GraphSAGE) \cite{hamilton2017inductive}
% The construction of the data set was done with the help of
% Open Graph Benchmark (OGB) \cite{hu2020open}. All models have been trained on a V100 GPU. We also provide the source code\footnote{The code is provided as supplementary material along with the paper. The link to the code is omitted to preserve anonymity.} for the model proposed in this paper.
% \subsection{Negative Sampling}
% \label{sec:neg_sampling}
% Most graphs used in GNNs do not provide ground truth negative edges, but the ST${_1}$, ST${_2}$, and ST${_3}$ datasets contain ground truth negative edges, which are based on the returned products. During the training part, the negative edges were sampled in different ways based on:
% \begin{enumerate}
%     \item their frequency (edge weight) using a multinomial distribution.
%     \item a uniform random distribution that samples from the ground-truth negative edges. 
%     \item a uniform random distribution which samples from all non-positive edges.
%     \item a ratio of all the aforementioned.
% \end{enumerate}

% \subsection{Parameters Setting}
% \label{sec:parameters}
For all experiments, we chose the following hyperparameter settings: dropout for the input node features inside the GNN layer under the GNNRel block $ \in \{0.3, 0.4, 0.5, 0.6, 0.7, 0.8\}$. We have used Adam optimizer \cite{kingma2014adam} with a learning rate set to $0.0001$ and the weight decay set to $0.00001$. For CS$_{i}$ datasets, the features are generated using learnable embeddings.
% By learnable embedding, we mean that we use a one-hot encoding representation for the nodes. % and then we apply a linear layer where its weights are representing the node. 
For the ST$_{i}$ datasets, we used the 2048 dimensions produced from ResNet-50 \cite{he2016deep}. For all datasets, we have concluded with an architecture $2$-$1$-$1$, which means that we use two GNN subblocks for the first GNNRel block and a single GNN subblock for the second and third blocks, respectively. The GNN in the first block has [$2048, 520$] layer size for the ST${_i}$ datasets and [$200,150$] for the CS$_{i}$ datasets. The layer size for the second block is set to [$250$] for the ST${_i}$ and [$100$] for the CS$_{i}$ datasets. The last block is set to a layer size equal to [$150$] for the ST${_i}$ and [$75$] for the CS$_{i}$ datasets. 

% The activation function for the GNN is decided to be ReLU \cite{agarap2018deep} as shown in Section \ref{sec:GOOD}. The embeddings generated from GNNRel blocks are aggregated and then fed to two different Feed Forward Networks, where the first one uses sigmoid activation on the aggregated embeddings for solving the link prediction task and the second one is using Softmax for solving the coefficient disentangling. 

To find the best set of parameters for our model in different experiments, we use the Ray\footnote{
\url{https://docs.ray.io/en/latest/tune/examples/tune-pytorch-lightning.html}
}
framework which performed hyperparameter searching using Hyperopt \cite{bergstra2013making} as backend. 
% For further details on the implementation, refer to Appendix \ref{sec:implementation_details}.
The parameters we performed the hyperparameter search are the dropout value, the choice of the aggregation function, the existence of the BN \cite{ioffe2015batch} layer, and the residual connections.

\section{Ablation studies and Model Analysis}
\label{sec:ablation_studies}
% \label{sec:impact_coef}

Revisiting the architecture Fig.~\ref{fig:arch_diagram}, we can see that our model consists of two key components, namely, GNNRel, which constructs the context-specific node embeddings, and MixAGG, which aggregates all context-specific embeddings. 

\begin{table}[tp]
    \caption{Impact of using random coefficients with Coefficient Disentangler as compared to equal contribution.}
    \centering
    \label{tab:prob_ablation}
  \resizebox{\columnwidth}{!}{%
    \begin{tabular}{lcccc}
% \hline
\toprule
\multirow{2}{*}{\textbf{ALGORITHM}} & \multicolumn{2}{c}{\textbf{ST$_1$}}  & \multicolumn{2}{c}{\textbf{CS$_2$}}  \\ 
% \cline{2-5}
& \textbf{Accuracy ($\%$)} & \textbf{ROC-AUC} & \textbf{Accuracy ($\%$)} & \textbf{ROC-AUC} \\
\midrule
GOOD & \textbf{71.33} & 0.6307 & 66.12  & \textbf{0.6990}  \\ 
% \hdashline
GOOD$_{\text{LC}}$ & 69.28 & \textbf{0.6335} & \textbf{63.75} & 0.6884 \\
% \hdashline
GOOD-coefficient-ablated  & 69.80 & 0.5944 & 62.41 & 0.6613 \\ 
% \hline
\bottomrule
\end{tabular}%
    }    
\end{table}
\textit{Impact of coefficients aggregation (MixAGG)} \label{sec:impact_coef}: \autoref{tab:prob_ablation} reports the relative performance, in terms of accuracy and ROC-AUC, for ST$_1$ and CS$_2$ data sets of different variants. All variations have been trained with BCE as their objective function, except GOOD, which used BCE in combination with MSLE. The MixAGG of the GOOD model used random coefficients, whereas the MixAGG of the GOOD$_{\text{LC}}$ model used learnable mixing coefficients. The MixAGG of the GOOD-coefficient-ablated model used fixed normalized coefficients during training and inference. 
As we can observe in \autoref{tab:prob_ablation}, GOOD, which disentangles the coefficients of the multiple relations in the final node embedding, performed better compared to the GOOD-coefficient-ablated model, which instead aggregates all embeddings with equal contribution without generalizing on the different values of the coefficients.  Furthermore, the GOOD$_{LC}$ model seems to generalize better than a MixAGG with fixed coefficients.
Therefore, our method for decomposing these different contributions is effective. 

\IEEEtriggeratref{39}
\bibliographystyle{IEEEtran}
\bibliography{reference.bib}
\end{document}